\documentclass[runningheads]{llncs}
\usepackage[T1]{fontenc}
\usepackage{amsmath} 
\usepackage{float}

\usepackage{graphicx}
\begin{document}
\title{Self-supervised structured object representation learning\thanks{This manuscript has been submitted to International Symposium on Visual Computing (ISVC) and has been under review since August 20, 2025.}}
\titlerunning{Structured Object Abstractions in SSL}
%
%
\author{Oussama Hadjerci\inst{1}\orcidID{0000-0003-4601-8778} \and
	Antoine Letienne\inst{1}\orcidID{0009-0008-6915-1938} \and
	Mohamed Abbas Hedjazi\inst{1}\orcidID{0000-0001-9492-3719}
	Adel Hafiane\inst{3}\orcidID{0000-0003-3185-9996}}

\authorrunning{O. Hadjerci et al.}

\institute{DASIA, Courbevoie, 92400, France.\\
	\email{\{oussama.hadjerci,antoine.letienne\}@dasia.ai}
	\and
	Université d'Orléans, INSA CVL, Bourges, 18000, France.\\
	\email{adel.hafiane@insa-cvl.fr}}

\maketitle              
\begin{abstract}
Self-supervised learning (SSL) has emerged as a powerful technique for learning visual representations. While recent SSL approaches achieve strong results in global image understanding, they are limited in capturing the structured representation in scenes. In this work, we propose a self-supervised approach that progressively builds structured visual representations by combining semantic grouping, instance level separation, and hierarchical structuring. Our approach, based on a novel ProtoScale module, captures visual elements across multiple spatial scales. Unlike common strategies like DINO that rely on random cropping and global embeddings, we preserve full scene context across augmented views to improve performance in dense prediction tasks. We validate our method on downstream object detection tasks using a combined subset of multiple datasets (COCO and UA-DETRAC). Experimental results show that our method learns object centric representations that enhance supervised object detection and outperform the state-of-the-art methods, even when trained with limited annotated data and fewer fine-tuning epochs.
\end{abstract}
\section{Introduction}
Large language models such as GPT~\cite{radford2018gpt}, BERT~\cite{devlin2019bert}, and LLaMA~\cite{touvron2023llama} have demonstrated remarkable performance across a wide range of tasks, largely due to their ability to learn rich representations directly from row data. A key factor behind their success is the use of self-supervised learning (SSL), where models are trained to predict parts of their input from other parts, without relying on human annotations. The computer vision community has been inspired by SSL paving the way for more robust, unbiased, and generalizable systems that can scale to open world data and complex scenarios.

SSL approaches such as DINO~\cite{caron2021dino}, iBOT~\cite{zhou2022ibot}, and BEiT~\cite{bao2021beit} have extended these ideas to image understanding. Most of these methods rely on large Transformer based architectures (VIT~\cite{dosovitskiy2021vit}) and learn global representations by solving auxiliary tasks such as predicting similarities between augmented views. These models have proven effective for image classification and matching, however, they often struggle to capture the compositional structure of objects in scenes.
MMethods like CutLER~\cite{wang2023cutler} typically identify a single dominant foreground region but struggle to isolate multiple objects without costly iterative training. Slot Attention~\cite{locatello2020slot} or SlotCon~\cite{wen2022slotcon} aim to produce structured scene representations through implicit grouping mechanisms, but lacks explicit hierarchical organization and instance level decomposition.

Current limitations are twofold. First, their design tends to produce representations of entire scenes, with limited capacity for structured object understanding. Yet an image is far more than a single label, it contains objects, parts, and spatial relationships. A meaningful visual representation should reflect this multi level structure and represent features at local, intermediate, and global scales. Second, fine-tuning these models for dense prediction tasks often demands substantial computational resources and extensive annotated data, making deployment challenging in resource constrained settings. 

We introduce an SSL approach based on a new module called ProtoScale designed to overcome these limitations by progressively constructing a structured visual representation that combines semantic abstraction, instance separation, and hierarchical organization. Our approach builds on the intuition that an effective representation should be able to: (i) group pixels based on semantic meaning, (ii) identify distinct parts of instances, and (iii) organize them coherently to represent complete objects. To test this hypothesis, we focus on the task of object detection, which offers a good balance between local and global scene complexity for interpretation. The objective is to capture complexity of visual scenes without relying on manual annotations, while supporting hierarchical object learning at multiple scales. Unlike state-of-the-art (SOTA) methods that rely on disruptive cropping strategies, we preserve the full scene context to ensure instance consistency while maintaining spatial structure. In addition, we adopt a lightweight hybrid encoder to reduce computational cost and improve feature scalability. We assess the representation learning ability of the proposed model by conducting downstream evaluations on the object detection task, using a combined subset of multiple datasets including COCO~\cite{lin2014coco} and UA-DETRAC~\cite{wen2020ua}. We challenge our method under conditions where annotations are limited and training does not rely on large scale computational resources. Unlike traditional downstream task pipelines based on Mask R-CNN, which rely heavily on post processing steps such as Non Maximum Suppression (NMS) and are often less efficient in real world applications, we adopt a DETR-like architecture~\cite{carion2020detr}, which provides a fully end to end formulation without NMS, and better aligns with our goal of evaluating structured, object centric representations in a simplified and deployable detection setting. Our method achieves strong performance on COCO and UA-DETRAC, even when trained for only $10$ epochs and with just $50$\% of the annotated training data.

In summary, the main contributions of this paper are as follows: (1) SSL approach that combines semantic grouping, instance level clustering, and hierarchical reasoning to build structured visual representations without any annotations. (2) ProtoScale, a modular grouping module operating at multiple spatial scales, capable of organizing pixels into instances and instances into structured groups. (3) Preserving full scene context across views by avoiding disruptive cropping, improving consistency in dense prediction tasks. (4) Leveraging a lightweight hybrid CNN/Transformer encoder, adapted from RT-DETR~\cite{zhao2024rtdetr}, to balance feature precision and reduce computational overhead. (5) Validation of the proposed approach through downstream object detection experiments by comparing it with SOTA methods from the literature and evaluating it under realistic constraints, using a small annotated dataset and a limited number of training epochs.

\section{Related work}

Most SSL methods in computer vision relies on contrastive learning, where the objective is to bring closer the representations of two different views of the same image. The approaches in SimCLR~\cite{chen2020simclr} and MoCo~\cite{he2020moco}, typically use two networks, one for the original image and the other with its transformed one, aim to maximize their similarity in latent space.

DINO~\cite{caron2021dino} extends this paradigm by introducing a teacher–student framework, where the teacher model stabilizes the training of the student via momentum updates. This framework leads to the spontaneous emergence of structured visual patterns such as foreground and background regions. Several variants such as  iBOT~\cite{zhou2022ibot}, and TEC~\cite{gao2022tec} further enrich this contrastive logic by incorporating multi view representations and dense patch level objectives. While these methods have driven significant progress in representation learning, they remain limited in their ability to structure objects in images. Their primary goal is to encode an entire image as a global vector, which is suitable for image level tasks like classification, but fall apart for dense prediction.

Some recent efforts have tackled the problem of object discovery without supervision by leveraging perceptual grouping cues, through clustering as in MOST~\cite{rambhatla2023most}, or an iterative supervised learning loop in the case of CutLER~\cite{wang2023cutler}. While effective at discovering salient regions, these approaches remain limited by the quality of the initial pseudo labels and often focus on a single dominant object per image.

In parallel, dense SSL methods have emerged, aiming to align features at the pixel or region level, like DetCon~\cite{henaff2021detcon}, and SoCo~\cite{wei2021soco}. These methods learn features that preserve spatial coherence and encourage local consistency across regions within an image. Other works use latent slots as intermediate representations to structure objects in scenes. Slot Attention~\cite{locatello2020slot} uses an iterative attention mechanism to assign pixels to abstract instances. SlotCon~\cite{wen2022slotcon} combines semantic grouping with contrastive learning to extract salient, object centric representations. Inspired by SlotCon, we extend its formulation by modifying the semantic grouping process and explicitly modeling hierarchical relationships between instances. 

Most recent SSL methods rely on pure Transformer based architectures, especially Vision Transformers (ViTs), which have shown strong performance on classification tasks. However, ViTs are computationally expensive in both memory and training time. Moreover, their patch based processing scheme fragments the input and removes local spatial continuity, limiting their ability to capture small or occluded objects. In this work, we adopt a CNN/Transformer hybrid encoder proposed by RT-DETR~\cite{zhao2024rtdetr} to reduce computational cost while preserving fine spatial cues, which is particularly beneficial for unsupervised dense prediction.

While recent SSL methods have made impressive progress, they remain incomplete when it comes to producing structured image representations. Contrastive methods yield effective global or local representation but lack grouping capabilities. Latent slot based models offer encouraging directions but still fall short in terms of hierarchy, modularity, and generalization. We addresses this gap by proposing a modular, hierarchical grouping mechanism based on shared visual prototypes, integrated into contrastive learning framework.

\section{Object-level contrastive learning}
\begin{figure}
\centering
\includegraphics[width=1\textwidth]{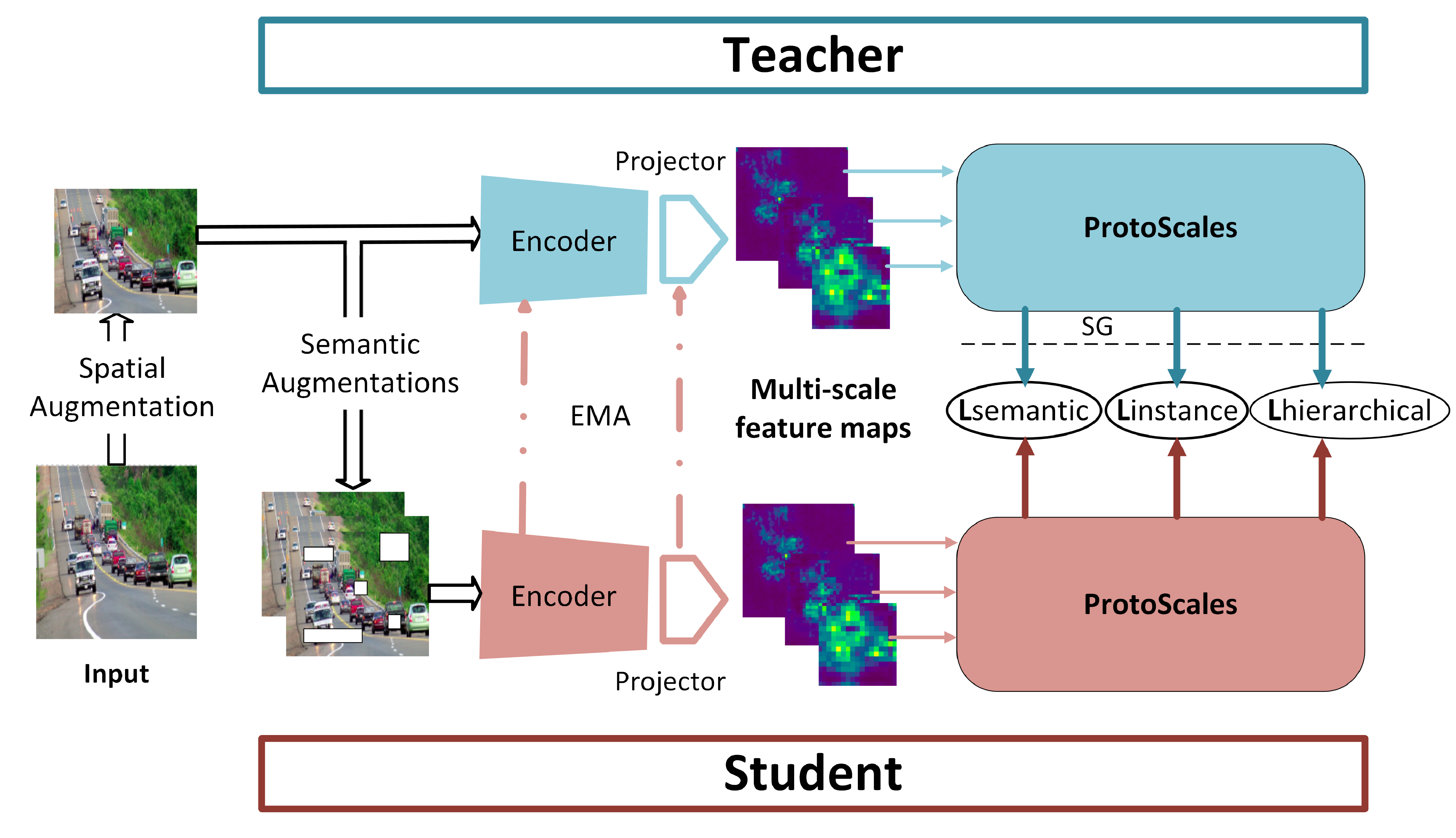}
\caption{Overview of the proposed training framework integrating multi-view learning, multi-scale feature encoding, and the ProtoScale grouping module. The teacher network (depicted in blue) is frozen (SG: stop-gradient), while the student network (depicted in red) is trainable.} \label{framework}
\end{figure}

The proposed method is based on the student–teacher framework (see Fig.~\ref{framework}), commonly used in self-supervised learning. Both networks share the same encoder architecture but differ in their input processing and training roles. Unlike SOTA methods, the student is fed with multiple semantically diverse yet spatially coherent augmented views of each input image, while the teacher receives a single global augmented view with stronger spatial transformations and serves as a stable label generator. Its parameters are updated using an exponential moving average (EMA) of the student weights. 
Training encourages the student predictions across all views to align with the teacher output using three loss terms, promoting view consistent grouping. These losses are tightly integrated with our proposed module, ProtoScale, which groups elements of the scene based on their semantic meaning, instance assignment, and hierarchical relations. This module favors a cross perspective understanding of object structure and enables scene dense parsing. To capture visual regularities at different levels of abstraction, both models share a multi scale encoder. This encoder extracts feature maps at three spatial resolutions, enabling ProtoScale to operate simultaneously at fine, intermediate, and global scales. This design helps the model discover small object parts, full object, and global scene in a unified feature representation.
\subsection{Multi-view learning}
Our approach relies on an asymmetric multi view learning strategy to encourage consistent grouping across appearance variations. The teacher processes a single globally augmented image obtained through a sequence of spatial transformations, including random cropping, zoom out, horizontal flipping, and photometric distortions. This image serves as the stable target for training and as the base input from which the student views are generated. In contrast, the student is exposed to four distinct views of the same input, sampled from a different transformation pipeline. These augmentations are designed to be spatially coherent, preserving object layout and relative positioning, while introducing sufficient appearance variability. Specifically, they include Gaussian blur, random spatial masking, and color jitter. Unlike SOTA methods that rely heavily on random cropping strategies which disrupt the spatial distribution of objects and degrade dense prediction performance, we adopt a strategy that preserves the full scene context across views. This design choice promotes instance consistency without compromising the underlying spatial structure.
\subsection{Hybrid CNN-Transformer pyramid encoder}
Unlike previous SSL methods, we employ a hybrid encoder architecture inspired by RT-DETR. The encoder processes multi scale features extracted from a ResNet~\cite{he2016resnet} backbone. The lowest resolution feature map is passed through a full self-attention transformer, capturing long range spatial dependencies in the image and providing a coarse indication of where the model should focus. This global attention serves as guidance to identify regions of interest in the image. The resulting low resolution attention is then progressively fused with intermediate and high resolution convolution features through a combination of top down and bottom up fusion paths. This stage relies on the local precision of CNN features to determine what is present in those regions. This efficient hybrid design enables accurate dense representation while reducing the computational cost associated with full resolution attention.
\vspace{-1em}
\begin{figure}
\centering
\includegraphics[width=0.8\textwidth]{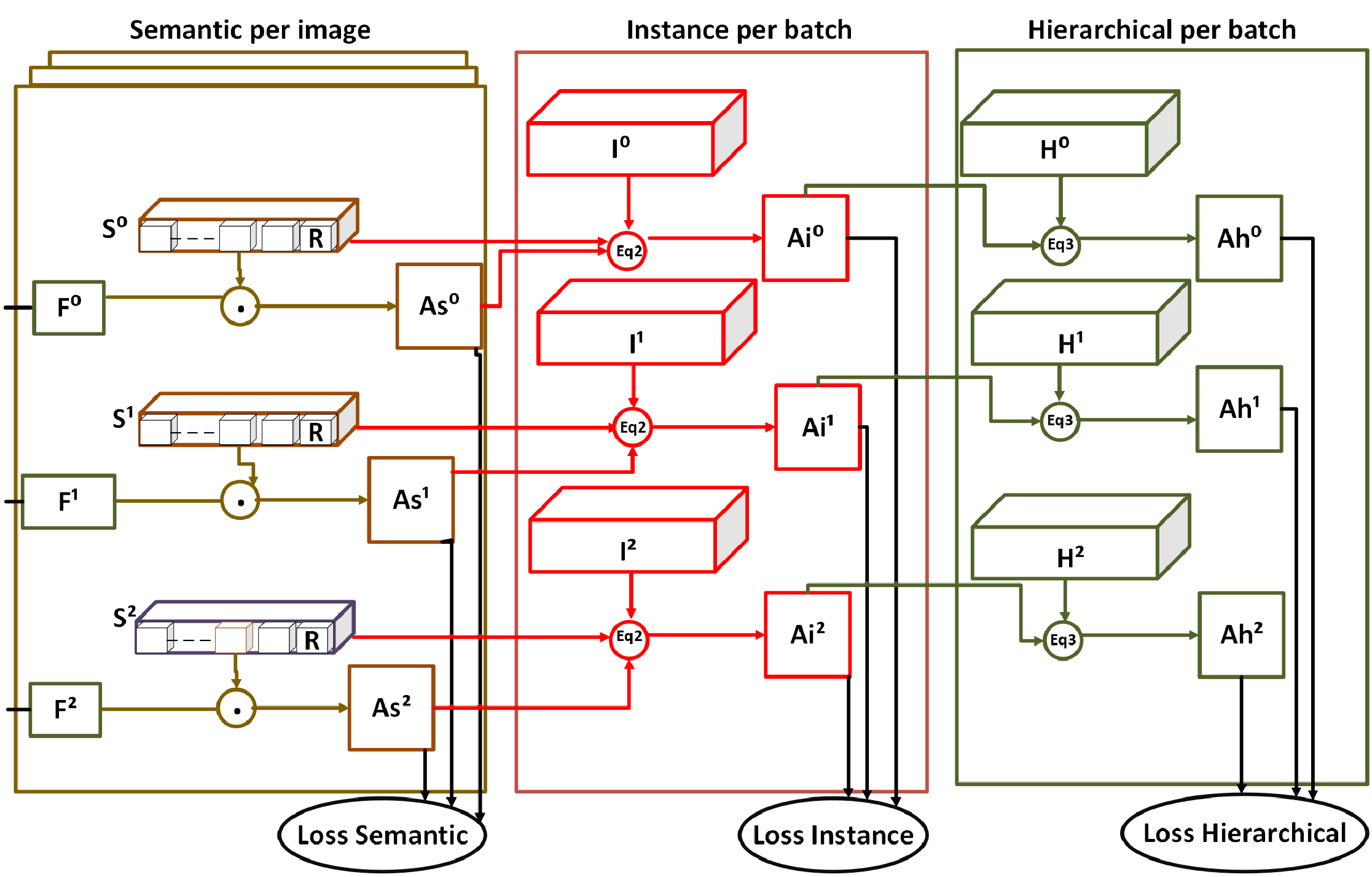}
\caption{Multi-level grouping process in ProtoScale module.} \label{protoScale}
\end{figure}
\vspace{-3em}
\subsection{ProtoScale: a multi-level grouping modules}
Once the features are extracted from the encoders, they are processed by our proposed grouping module, ProtoScale (Fig.~\ref{protoScale}). This module consists of three sequential components. First, a semantic clustering stage is applied independently to each feature map, where spatial attention maps are generated using learnable prototypes (cluster centers). Second, the resulting semantic prototypes from the entire batch are aggregated into a shared set of instance level prototypes, enabling consistent grouping of object-like regions across different samples. Finally, a lightweight hierarchical module infers relationships between these instances, producing a hierarchical grouping that captures object–part structures at the batch level.

\subsubsection{Semantic grouping}
At each output scale $k$ of the encoder, spatial features are projected and fed into a semantic grouping module based on prototypes attention. Inspired by SlotCon, we compute dot product attention between $256$
learnable semantic prototypes $S_k \in \mathcal{R}^{256}$ and spatial features \( F_k \) as follows:
\begin{equation}
    As_{k} = \text{softmax} \left( \frac{S_k^\top \cdot F_k}{\tau} \right),
\end{equation}
with $\tau > 0$ temperature parameter that control the sharpness of the output distribution. Each prototype \( S_k \) aggregates features through weighted attention to form semantic queries. These attention maps \( As_k \) highlight spatial groupings based on shared appearance features.

Unlike SlotCon, we integrate this grouping into a dense prediction pipeline by applying a supervised loss between the student and teacher attention maps at multiple scales. Furthermore, we introduce a centered Gaussian prior ($\mu=0.5, \sigma=0.7$) applied to the attention maps to reduce collapse and mitigate the bias toward generic image representations, encouraging focus on local structure instead. Inspired by the work of \cite{jiang2025vision}, we also incorporate a set of auxiliary prototypes $R$ (see Fig.~\ref{protoScale}). designed to absorb noisy activations, these are excluded from the loss to stabilize learning. This setup enables the model to learn interpretable, multi level semantic decompositions.

\subsubsection{Instance grouping via shared latent prototypes}

While semantic grouping isolates regions, it fail in crowded or occluded scenes. We address this by introducing a differentiable instance grouping module based on shared latent prototypes $I_k$, learned across the batch.

Each slot \( S_k \) is softly assigned to the $32$ learnable parameter ($I_k \in \mathcal{R}^{256}$). The resulting assignment weights are then used to scale the corresponding attention map \( As_k \) before summation over all semantic prototypes ($Np$) :
\begin{equation}
    Ai_{k} =  \sum_{i=1}^{Np} 
\text{softmax}_i\left( \frac{ \left\| S_k \right\| \cdot \left\| I_k \right\| }{\tau} \right) \, As_k,
\end{equation}
with $\tau > 0$ temperature parameter. These instance attention reflect coherent object level regions. The sub-module is trained with consistency and sparsity regularization, encouraging specialization and spatial diversity among instances prototypes.

\subsubsection{Hierarchical object understanding through prototype affinities}
Beyond instances level attentions, ProtoScale also capture object relationships by modeling the affinity between instance prototypes. For each scale $k$, a relation matrix ($H_k = \sigma ( \text{MLP} \left( I_k) \right)$) between instances prototypes \(I_k\) is predicted using a shallow multilayer perceptron ($MLP$), followed by sigmoid activation ($\sigma$). This matrix is used to define hierarchical attention ($Ah_k = H_k \cdot Ai_{k}$). By thresholding low affinity pairs at $0.5$, the model learns to merge overlapping or related instances, capturing part-whole and object group structures. This module enriches the learned representation with scene level structure and abstraction.

\subsection{Self-supervised objective}
The student is trained to match the teacher outputs using a weighted objective composed of three terms, a semantic loss, an instance level loss, and a hierarchical loss. The overall loss function is defined as : $
\mathcal{L}_{\text{total}} = \lambda_{\text{sem}} \mathcal{L}_{\text{sem}} + \lambda_{\text{inst}} \mathcal{L}_{\text{inst}} + \lambda_{\text{hier}} \mathcal{L}_{\text{hier}}$, where $(\lambda_{\text{sem}}, \lambda_{\text{inst}}, \lambda_{\text{hier}}) = (1, 2, 1)$ yields the best results in our experiments.
Each loss includes a KL divergence term computed between the corresponding attention maps of the student and the teacher. The KL divergence was chosen to encourage the student model to match the entire probability distribution rather than focusing only on the highest-probability regions. All modules operate at three spatial scales, enabling rich object-centric reasoning across fine, intermediate, and global resolutions.

\section{Experimental setting}
In this section, we first describe the baselines used in our evaluations, then present the datasets, and finally explain the evaluation setup.
\subsection{Baselines}
Self-supervised encoders are typically evaluated through downstream fine-tuning on detection or segmentation tasks, often using Mask R-CNN. However, in this work, we adopt the RT-DETR framework, which offers an end to end detection pipeline without relying on post processing steps such as non maximum suppression. Moreover, RT-DETR achieves significantly better performance compared to earlier methods like Mask R-CNN. Unfortunately, RT-DETR does not support instance segmentation, which prevents us from evaluating segmentation performance. Therefore, we focus exclusively on object detection as the downstream task. We select a set of strong SSL baselines from the literature: DINOv1~\cite{caron2021dino}, DINOv2~\cite{oquab2024dinov2}, iBOT~\cite{zhou2022ibot}, TEC~\cite{gao2022tec} and SlotCon~\cite{wen2022slotcon}. For each method, we use the encoder backbone and attach it with a shared projection module that outputs multi scale features. These encoders are integrated into our student–teacher method for a fair comparison under the same training configurations. We also report results for the fully supervised RT-DETR~\cite{zhao2024rtdetr} model as an reference.
\subsection{Dataset}
We conduct experiments on two datasets selected for their complementary characteristics offering high scene diversity and high object density, respectively for COCO and UA-DETRAC. To simulate real world scenarios, we redefine the original class labels into a fine-grained taxonomy of 17 classes. We extract subsets from each dataset by selecting only images containing at least two of these classes, resulting in $39429$ training and $4500$ validation images from COCO, and $10539$ training and $1055$ validation images from UA-DETRAC. For a combined total of $50022$ training and $5555$ validation images.
\subsection{Evaluation setups}
\begin{figure}
    \centering
    \includegraphics[width=.95\linewidth]{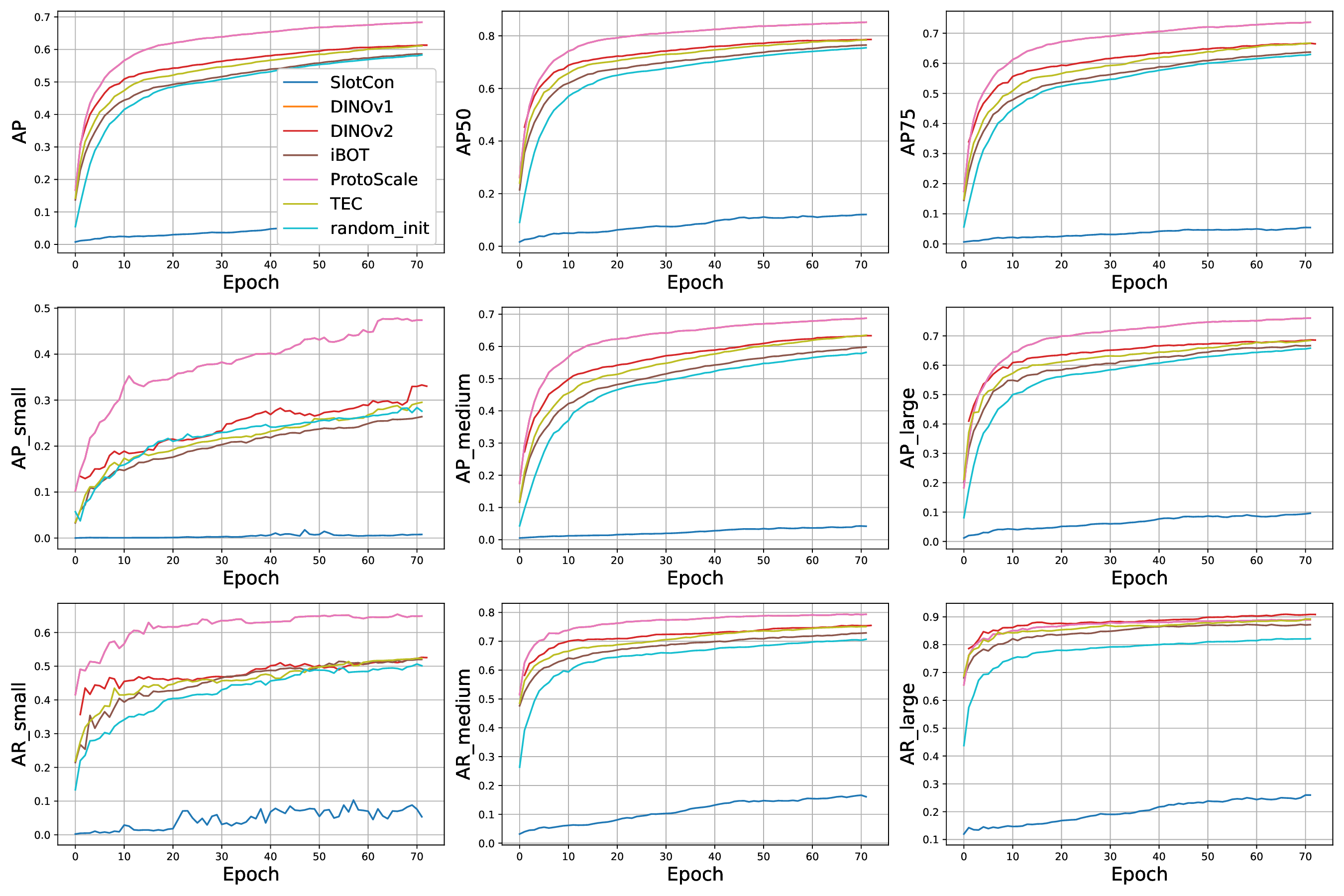}
    \caption{Main Transfer Results with SSL Pre-training over Epochs}
    \label{expepochs}
\end{figure}

The core principle of SSL is to enable the fine-tuning of downstream models with limited labeled data and within a small number of epochs, while still achieving strong generalization. Following standard practice in SSL, we assess the representation capability of the pre-trained model by using it as the backbone for a downstream detection task. Instead of Mask R-CNN, we adopt RT-DETR and fine-tune all layers end to end using different SSL pretrained backbones. To simulate a realistic low resource scenario, we begin by fine-tune the detection model for a varying number of epochs and compare its performance with related methods (see Fig.~\ref{expepochs}). We then evaluate the impact of using different portions of the training data (see Tab.~\ref{expproportion}), while always testing on the full validation set. These experiments reflect challenging conditions in which annotated data and computation resources are limited, typical of real world environments where access to high computing infrastructure and large annotated datasets for fine-tuning is often unavailable. All experiments are conducted on dual NVIDIA RTX A6000.
\vspace{-2em}
\vspace{-3.5em} 
\section{Results and discussion}
In this section, we evaluate the performance of our SSL model through a downstream object detection task. We compare ProtoScale with several SOTA SSL methods on a combined subset of multiple datasets. Performance is reported using the standard COCO detection metrics. $AP$ and $AR$ denote the mean Average Precision and mean Average Recall, respectively, over IoU thresholds $[0.50:0.95]$. $AP_{50}$ and $AP_{75}$ correspond to $AP$ at $IoU = 0.50$ and $IoU = 0.75$, respectively. Results are also reported by object scale: small ($AP_s$, $<32^2$ px), medium ($AP_m$, $32^2 \leq$ area $<96^2$ px), and large ($AP_l$, $\geq96^2$ px). $AR$ is defined analogously under the same settings.

\subsection{Sensitivity to fine-tuning epochs}
In Fig.~\ref{expepochs}, we present the main results using models pre-trained on COCO~\cite{lin2014coco}, ImageNet-1k~\cite{deng2009imagenet} and LVD-142M~\cite{oquab2024dinov2} datasets. Specifically, we include SlotCon pre-trained for 800 epochs on COCO, DINOv1 with ViT Small trained for 300 epochs on ImageNet-1K, DINOv2 with ViT Small trained for 300 epochs on LVD-142M, iBOT with ViT Small trained for 800 epochs on ImageNet-1K, TEC with ViT-base trained for 800 epochs on ImageNet-1K, and Protoscale pre-trained for 35 epochs on COCO. We also include the RT-DETR baseline with random initialization as a supervised reference. All methods are fine-tuned on the combined dataset comprising COCO and UA-DETRAC.
ProtoScale significantly outperforms current SOTA SSL approaches, as well as the supervised RT-DETR with random initialization, consistently achieving higher performance even compared to models pre-trained on large scale datasets such as ImageNet-1K and LVD-142M. Notably, our method demonstrates strong performance on small and medium objects, even with fewer fine-tuning epochs. For example, SlotCon requires up to 800 epochs of pre-training to achieve competitive results. However, DINOv2 slightly outperforms our method on large objects, likely due to the strong generalization capabilities of transformer based architectures trained on a massive dataset of 142 million image.
\vspace{-1em}
\begin{table}[H]
\centering
\caption{Detection performance under varying proportions of labeled training data with 10 fine-tuning epochs.}
\label{expproportion}
\begin{tabular}{|l|l|l|lll|l|lll|}
\hline
Training set & Methods & $AP$& $AP_s$ & $AP_m$ & $AP_l$ & $AR$ &  $AR_s$ & $AR_m$ & $AR_l$\\
\hline
10\% & DINOv2 &   \textbf{20.1} &  $6.5$  & $17.$2 &  \textbf{29.58} & $57.9$ & $31.1$ & $52.5$ &  \textbf{73.0}  \\
10\% & ProtoScale &   $19.9$ &    \textbf{10.4} &  \textbf{17.8} & $22.5$ &  \textbf{58.2} &  \textbf{36.7} &  \textbf{53.3} & $70.2$  \\
\hline
30\% & DINOv2 &   $37.5$ &   $13.14$ & $33.4$ & \textbf{50.1} & $63.6$ & $39.4$ & $60.0$ & $78.1$  \\
30\% & ProtoScale &    \textbf{41.0} &   \textbf{21.1} & \textbf{39.3} & $48.2$ & \textbf{68.4} & \textbf{52.1} & \textbf{66.5} & \textbf{79.2}  \\
\hline
50\% & DINOv2 &   $50.7$ &   $17.9$ & $46.6$ & \textbf{65.8} & $68.9$ & $41.9$ & $66.0$ & $83.8$  \\
50\% & ProtoScale &    \textbf{57.1} &   \textbf{31.2} & \textbf{56.6} & $65.3$ & \textbf{75.2} & \textbf{59.7} & \textbf{74.2} & \textbf{85.2}  \\
\hline
70\% & DINOv2  &   $50.6$ &   $18.11$ & $46.8$ & $65.15$ & $69.10$ & $44.26$ & $66.0$ & $83.9$  \\
70\% & ProtoScale &    \textbf{58.7} &   \textbf{31.0} & \textbf{57.8} & \textbf{68.0} & \textbf{76.5} & \textbf{61.4} & \textbf{75.9} & \textbf{86.6}  \\ 
\hline
100\% & DINOv2 &   $63.7$ &   $36.2$ & $66.6$ & $70.0$ & $76.6$ & $58.4$ & $77.6$ & \textbf{91.5}  \\ 
100\% & ProtoScale &    \textbf{68.3} &   \textbf{47.4} & \textbf{68.7} & \textbf{76.1} & \textbf{79.8} & \textbf{64.8} & \textbf{79.3} & $88.9$  \\
\hline
\end{tabular}
\end{table}
\vspace{-3em}
\subsection{Influence of labeled data availability}
In Tab.~\ref{expproportion}, we further analyze the impact of limited labeled data on fine-tuning performance for the object detection task. We select different proportions of the training set (10\%, 30\%, 50\% and 70\%) to simulate low resource scenarios. We then compare the two best performing methods from the previous experiment (ProtoScale and DINOv2) to evaluate their robustness under varying data availability. Although the overall performance decreases compared to fine-tuning on 100\% of the training data, the results demonstrate that our approach requires less labeled data to achieve competitive performance. Notably, ProtoScale consistently outperforms DINOv2 on small and medium objects, while DINOv2 performs slightly better on large objects. Although the difference remains small and the DINOv2 encoder (ViT Small) contains approximately 21 million parameters, whereas our model uses only 12 million.

\section{Conclusion}
This work present a SSL approach that constructs structured visual representations through semantic grouping, instance level clustering, and hierarchical reasoning. Our proposed module, ProtoScale, operates at multiple spatial scales to organize scenes into semantically coherent and object centric components. By avoiding disruptive cropping and preserving full scene context across views, our method enhances consistency in dense prediction tasks. We validated our approach through downstream object detection experiments on combined subsets of COCO and UA-DETRAC, demonstrating strong generalization under realistic low resource constraints. ProtoScale consistently outperforms SOTA SSL methods, including those trained on large scale datasets, while using fewer parameters and requiring fewer training epochs. Future work will investigate the effectiveness of our method on larger datasets and explore its applicability to other dense prediction tasks. We also aim to design a SSL decoder to enhance the interpretability of the learned representations in a human-understandable manner.

\bibliographystyle{splncs04}
\bibliography{bibliography}
\end{document}